\documentclass[letterpaper, 10 pt, conference]{ieeeconf}  

\IEEEoverridecommandlockouts                              

\usepackage{graphics} 
\usepackage[pdftex]{graphicx}   
\usepackage{subcaption}         
\usepackage{epsfig} 
\usepackage{mathptmx} 
\usepackage{times} 
\usepackage{amsmath} 
\usepackage{amssymb}  
\usepackage{array} 

\title{\LARGE \bf

A Modular Pneumatic Soft Gripper Design for Aerial Grasping and Landing*
}

\author{Hiu Ching Cheung$^{1}$, Ching-Wei Chang$^{2}$, Bailun Jiang$^{1}$, Chih-Yung Wen$^{1}$ and Henry K. Chu$^{3}$
\thanks{*This work was supported by the Research Centre of Unmanned Autonomous Systems (RCUAS) [P0046487], The Hong Kong Polytechnic University}
\thanks{$^{1}$Hiu Ching Cheung, Bailun Jiang, and Chih-Yung Wen are with Department of Aeronautical and Aviation Engineering, The Hong Kong Polytechnic University,
        Hung Hom, Kowloon, Hong Kong
        {\tt\small athena-hiu-ching.cheung@connect.polyu.hk}, {\tt\small bailun-robin.jiang@connect.polyu.hk}, {\tt\small chihyung.wen@polyu.edu.hk}}%
\thanks{$^{2}$Ching-Wei Chang is with Hong Kong Center for Construction Robotics,
        , Hong Kong
        {\tt\small ccw@ust.hk}}%
\thanks{$^{3}$Henry K. Chu is with Department of Mechanical Engineering, The Hong Kong Polytechnic University,
        Hung Hom, Kowloon, Hong Kong
        {\tt\small henry.chu@polyu.edu.hk}}%
}

\begin{document}

\maketitle
\thispagestyle{empty}
\pagestyle{empty}

\begin{abstract}
Aerial robots have garnered significant attention due to their potential applications in various industries, such as inspection, search and rescue, and drone delivery. Successful missions often depend on the ability of these robots to grasp and land effectively. This paper presents a novel modular soft gripper design tailored explicitly for aerial grasping and landing operations. The proposed modular pneumatic soft gripper incorporates a feed-forward proportional controller to regulate pressure, enabling compliant gripping capabilities. The modular connectors of the soft fingers offer two configurations for the 4-tip soft gripper, H-base (cylindrical) and X-base (spherical), allowing adaptability to different target objects. Additionally, the gripper can serve as a soft landing gear when deflated, eliminating the need for an extra landing gear. This design reduces weight, simplifies aerial manipulation control, and enhances flight efficiency. We demonstrate the efficacy of indoor aerial grasping and achieve a maximum payload of 217 g using the proposed soft aerial vehicle and its H-base pneumatic soft gripper (808 g). 
\end{abstract}

\section{INTRODUCTION}
\subsection{Motivation}
\label{subsec::motivation}
Aerial grasping \cite{meng2022aerial} is a rapidly growing research field, particularly in the context of unmanned aerial vehicles (UAVs) used for drone delivery such as food delivery \cite{muchiri2022review}, search and rescue \cite{yeong2015review}, environmental cleaning \cite{qi2017design}, and medical goods delivery \cite{thiels2015use}. Ensuring the safety and efficiency of goods during drone delivery is critical. Hence, we aim to address these concerns by proposing a novel solution: a soft aerial vehicle (SAV) equipped with a modular soft pneumatic gripper, which has received limited attention in previous studies. 
\begin{figure}[t]
    \centering
    \includegraphics[scale = 0.30]{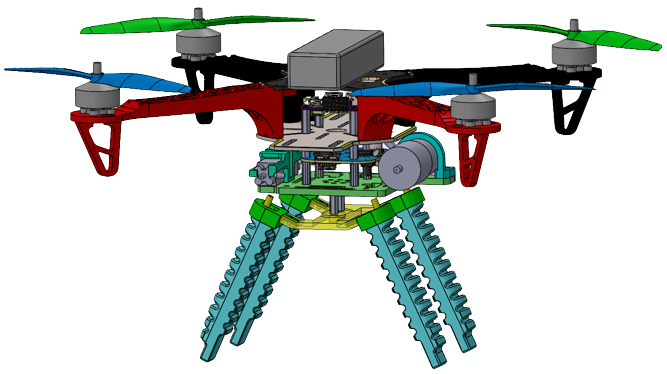}
    \caption{Illustration of the proposed modular pneumatic soft gripper with a conventional drone.}
    \label{fig::f330_sav}
\end{figure}
Soft grippers offer a promising alternative to rigid grippers due to their simplified control systems, reduced complexity, and excellent force absorption capabilities. To leverage the principles of soft robotics, these lightweight and controllable mechanisms provide stability during grasping by dampening impact forces. Their flexibility allows for a broader range of object dimensions, increasing grasping tolerance. Positioned beneath the center of gravity of the SAV, the soft gripper can inflate and deflate to securely grasp objects, improving grasping efficiency and reducing mission time. Moreover, replacing the rigid landing gear with the proposed soft gripper as the landing mechanism enhances the capability of aerial manipulation for the SAV \cite{fishman2021dynamic}. It eliminates the need for additional actuators and simplifies the control system by eliminating the complicated operations of the landing gear adjustment during takeoff, landing, and grasping. Furthermore, the modular design of the gripper enables configuration changes to accommodate various object sizes and shapes, for instance, considering the diverse shapes of food packages prevalent in food delivery scenarios.


\subsection{Related Work}
\label{sec::softgripper}
Inspired by biological systems like birds \cite{zhu2022perching}, soft grippers have demonstrated the ability to mitigate contact forces and compensate for grasping inaccuracies, exemplifying the concept of morphological computation, which combines passive mechanical components with explicit control \cite{rus2015design}. To leverage these advantages, some researchers have developed a bio-inspired tendon-actuated soft gripper, as the bending capability of the gripper can be easily controlled by manipulating the attached tendon \cite{king2018design, manti2015bioinspired, hassan2015design}. In the context of perching, Ramon et al. \cite{ramon2019autonomous} have designed a soft landing gear system that allows an aerial drone to autonomously perch on pipes in industrial environments, providing robust attachment using nylon strands and offering a modular design for gripper replacement. This soft gripper has demonstrated successful closure around pipes in outdoor experiments, enabling the quadrotor to perform well even in uncertain and windy conditions despite imperfect centering or fastening. Furthermore, Fishman et al. \cite{fishman2021control, fishman2021dynamic} have developed a tendon-actuated soft gripper for a soft drone inspired by the dexterity of human fingers. Their approach eliminates the need for gripper state measurement, operating the soft gripper in an open-loop manner to grasp objects of unknown shape on unknown surfaces, relying on the centroid of the objects.

The fabrication process for the tendon-actuated soft gripper involves the insertion of tubes inside the soft fingers to accommodate the passage of tendons \cite{peng2021aecom, chien2023design}. In contrast, a pneumatic soft gripper requires only a sufficient air chamber in each soft finger for bending \cite{min2023bayesian}. Moreover, the location of the tendon-actuated gripper is constrained by the position of the driven motor. Nevertheless, soft pneumatic grippers offer greater configurational flexibility due to the flexibility of the air tube. The air tube can quickly connect the pneumatic soft gripper with the pumps and valves. Related work also explores the potential of pneumatic soft systems in aerial grasping. Shtarbanov \cite{shtarbanov2021flowio} developed FlowIO, a wearable and modular pneumatic platform for soft robotics, which showcased the viability of pneumatic systems in miniaturized applications that are cost-effective for UAVs engaged in aerial grasping. Ping et al. \cite{pingaerial} integrated a soft pneumatic gripper beneath a conventional drone, similar to our design. However, their focus does not explicitly revolve around autonomous aerial grasping tasks or alleviating the need for additional rigid landing gear. The soft gripper in their work is controlled by a single command: pneumatic pressure. They claim that the installation of the entire Grasping Unit has minimal impact on the dynamics of the quadrotor. Since their proposed soft gripper does not stick out by adding extra degrees of freedom (DOF) to facilitate grasping, the rigid landing gear of their UAV limits the grasping altitude. A 2 DOF robot arm with a pneumatic soft gripper is installed under a traditional UAV with the rigid landing gear in \cite{sarkar2022development}. Compared to our proposed SAV, extra mechanisms and control algorithms in the work of Sarkar et al. \cite{sarkar2022development} are needed to ensure the manipulator can reach its home and pick positions. Also, their fixed, rigid landing gear increases the net weight of the UAV, which is over 3 kg in total. 

\subsection{Contribution}
\label{subsec::projectplanning}
The proposed SAV represents a significant advancement, capable of aerial grasping and landing using a lightweight modular soft pneumatic gripper. The pneumatic soft gripper, incorporating a feed-forward proportional controller for precise pressure regulation, also serves the dual purpose of functioning as a soft landing gear for the SAV during deflation. This innovative design eliminates the need for additional landing gear and allows the gripper to grasp various objects by its two configurations, resulting in a more streamlined and efficient aerial grasping system. Our contributions are summarized as follows.

\begin{itemize}
    \item Integrating modular connectors reduces the configuration and fabrication time for the soft gripper fingers. With just four fingers, the gripper can be arranged into two configurations, providing a versatile range of options for grasping target objects based on their shapes. This modularity enhances the adaptability of the gripper to grasp various objects during aerial operations. 
    \item The dual capability of the soft gripper to serve as both a grasping mechanism and a landing gear. The deflation of the soft gripper enables it to function as a soft landing gear, obviating the need for an additional rigid landing gear to reduce the weight of the SAV. This weight reduction enhances flight efficiency and simplifies the complexity associated with aerial operation control. 
    \item The proposed lightweight, soft gripper with a traditional quadrotor, which weighs merely 808 g (with its H-base gripper), successfully grasps a cylindrical plastic container weighing 217 g while in mid-air. This payload-to-weight ratio surpasses those observed in previous investigations \cite{fishman2021dynamic, pingaerial, sarkar2022development}.  
\end{itemize}


\section{MODULAR SOFT GRIPPER DESIGN}
\label{sec::soft_gripper_design_and_control}
\subsection{Mechanical Design}
\label{subsec::medesign}

\begin{figure}[t]
    \centering
    \includegraphics[scale = 0.28]{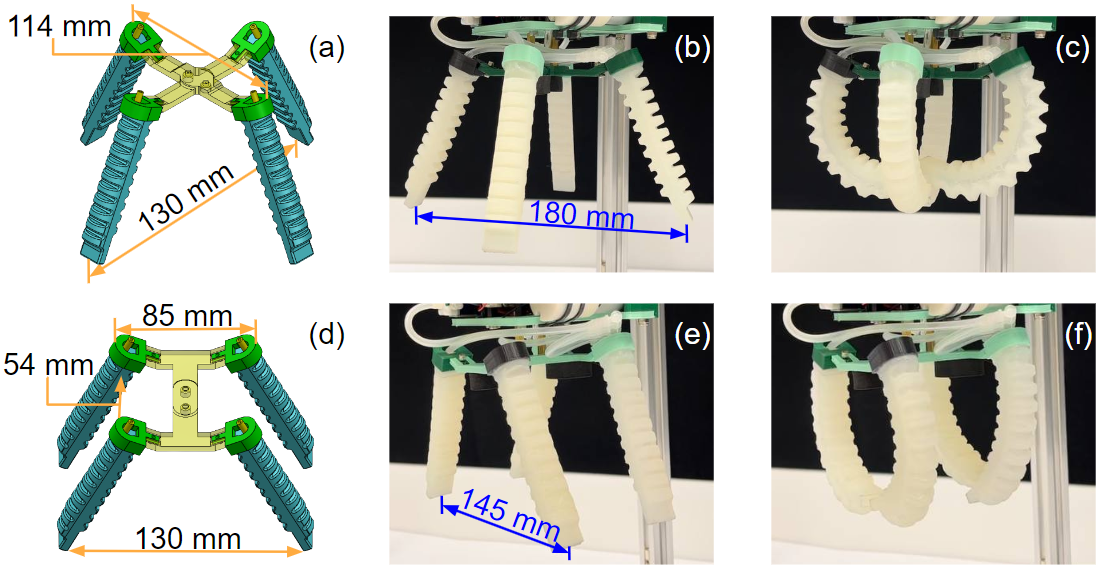}
    \caption{Dimensions of the X-base (spherical) soft gripper when it is (a) initially opened, (b) fully opened, and (c) fully closed. And the dimensions of the H-base (cylindrical) soft gripper when it is (d) initially opened, (e) fully opened, and (f) fully closed.}
    \label{fig::differentbasesofsoftgripper}
\end{figure}

The proposed soft fingers exhibit dimensions of 100 mm in length and 15 mm in width. These fingers can be affixed to multiple bases at a 25$^{\circ}$ inclination angle, giving the gripper increased grasping tolerance and diverse options for target objects. Grasping tolerance refers to the available dimension for grasping when the gripper is deflated, precisely the distance between the fingertips when the gripper is fully opened in Fig. \ref{fig::differentbasesofsoftgripper}(b) and Fig. \ref{fig::differentbasesofsoftgripper}(e), respectively. The diagonal distance of the fully opened X-base (spherical) soft gripper is 180 mm, while the tip-to-tip distance of the fully opened H-base (cylindrical) soft gripper is 145 mm. The dimensions of the 2 bases are shown in Fig. \ref{fig::differentbasesofsoftgripper}(a) and Fig. \ref{fig::differentbasesofsoftgripper}(d). When the complete closing of the soft gripper is undergoing, all tips of the soft fingers of the X-base soft gripper touch each other, while the tips of the two pairs of the 2-tip soft fingers of the H-base counterpart touch each other too (see Fig. \ref{fig::differentbasesofsoftgripper}(c) and Fig. \ref{fig::differentbasesofsoftgripper}(f)). 

The X-base and H-base configurations necessitate using four soft fingers to grasp target objects, as discussed in prior works \cite{pingaerial, tawk20223d}. Each finger of the soft gripper is equipped with a single air chamber. These air chambers are interconnected and equipped with one micro-pump and two solenoid valves. The weights of the H-base and X-base soft grippers are 106 g and 110 g (excluding the electronics), respectively. The modular connectors of the soft fingers have been designed to simplify the attachment of the soft fingers to various bases. Sliding the connector and securing it with a screw allows each soft finger to connect to different bases to grasp different objects quickly. This approach reduces the required number of soft fingers to a maximum of four, significantly reducing assembly time. 

\subsection{Fabrication}
\label{subsec::fabrication}

\begin{figure}[t]
      \centering
      \includegraphics[scale=0.3]{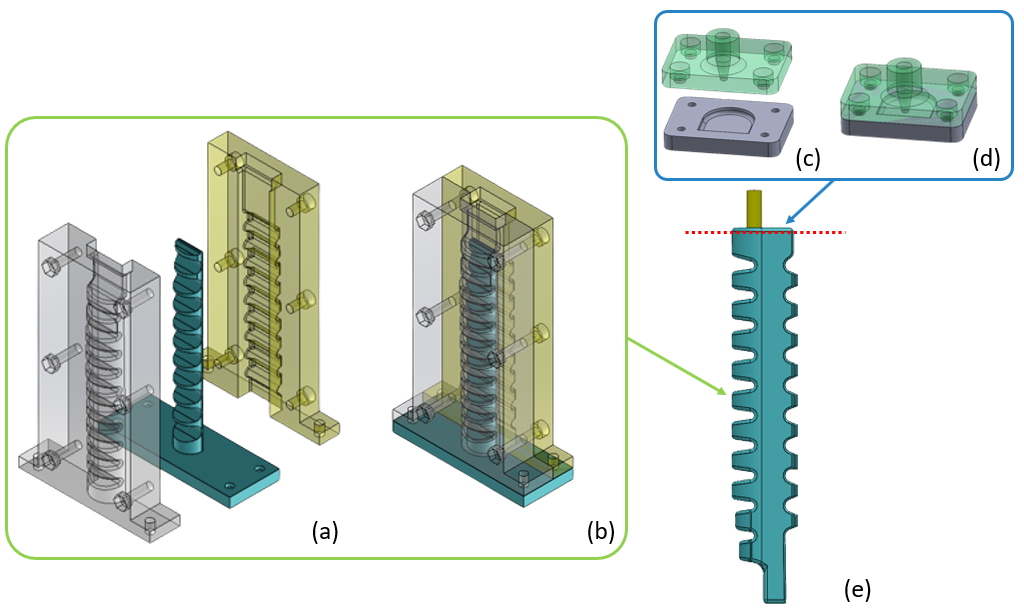}
      \caption{The exploded view (a) and assembly (b) of the mold of the soft finger’s main body. The exploded view (c) and assembly (d) of the mold of the soft finger’s cover. (e) The side view of a soft gripper.}
      \label{fig::mold}
\end{figure}
 
The soft fingers are molded using silicone rubber with a proposed shore hardness of 30 A. For manufacturing convenience, all the molds, soft finger connectors, and bases are 3D-printed using Polylactic Acid (PLA). Fig. \ref{fig::mold} illustrates the mold design for a soft finger, which consists of two parts. One part forms the main body of the soft finger, including its air chamber, while the other part produces the cover of the soft finger. This approach allows for efficient production and assembly of the soft fingers.
 
Referring to \cite{zhang2017modular}, the main steps of the soft finger fabrication include assembling the molds, applying mold release spray, pouring liquid silicone rubber, cutting holes in the covers, inserting and securing silicone tubes, and joining the main body and covers using silicone gel. The silicone tubes are not sealed by silicone gel, and they are only connected with the tubing connector by friction to enable the quick interchange between H-base and X-base.

Only the mold for making the air chamber is disposable because it will be twisted and bent during removal from the soft finger. The remaining parts of the mold can be reused for future fabrication processes.

\subsection{Electronics}
\label{subsec::electronics}

\begin{figure}[t]
      \centering
      \includegraphics[scale=0.30]{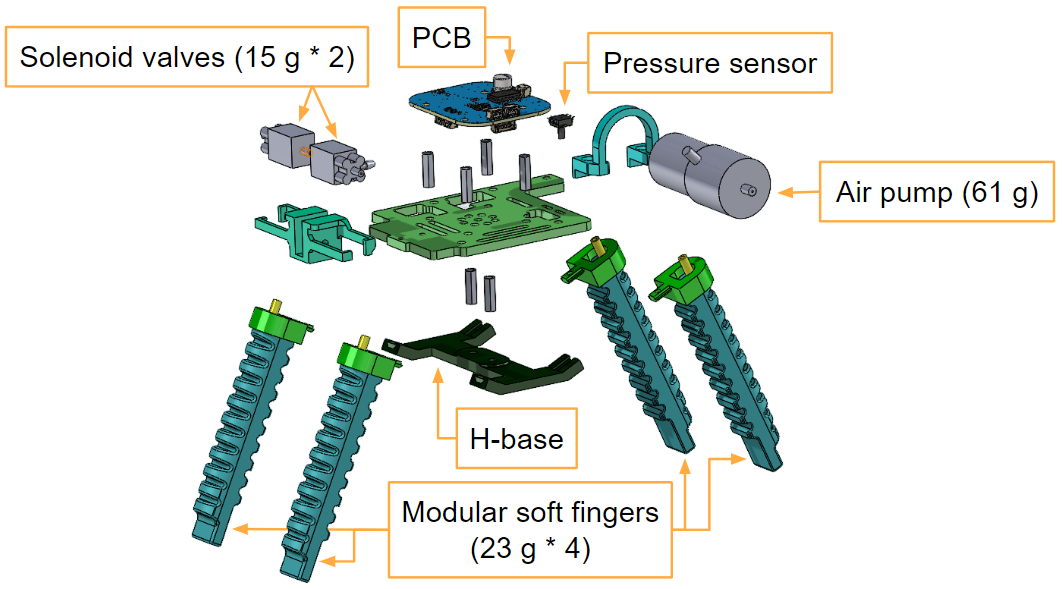}
      \caption{The exploded CAD view of the soft gripper (including weight information).}
      \label{fig::exploded_cad_gripper}
   \end{figure}

The main electronic components of the soft gripper include the microcontroller, air pump, solenoid valves, and pressure sensor. Control of the gripper is managed by an ESP32-S3 microcontroller, which receives a Pulse-width Modulation (PWM) signal from the flight controller for regulating the air pump and valve operations. The soft gripper is equipped with a single air pump that facilitates both inflation and deflation, while two valves serve as switches to control the airflow for inflation or deflation simultaneously. The air pump has a pressure range of [-60, 120] kPa, and a pressure sensor with a range of [-100, 300] kPa is incorporated to detect the internal airflow pressure within the gripper and provide feedback to the microcontroller. To meet all the requirements of the proposed soft gripper, except for the air pump and the valves, all the electronic components are integrated onto a customized Printed Circuit Board (PCB), as shown in Fig. \ref{fig::exploded_cad_gripper}. The total weight of the electronic components is 146 g.

In order to enhance the efficiency of the SAV by reducing its overall weight, the soft gripper does not require an additional power source when installed beneath a traditional quadrotor. This minimizes energy requirements, enabling the soft gripper to operate efficiently with the existing quadrotor power system. Hence, the payload capacity of the SAV can be maximized.

\section{MODULAR SOFT GRIPPER CONTROL}
\label{sec::control_of_soft_gripper}
\subsection{Airflow Control System}
\label{subsubsec::airflowcontrolsystem}

\begin{figure}[t]
      \centering
      \includegraphics[scale=0.35]{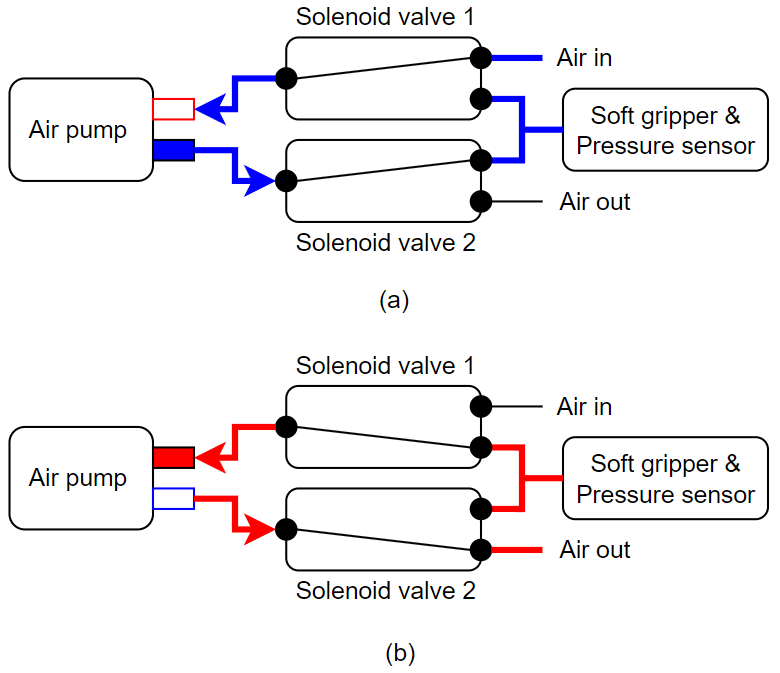}
      \caption{Airflow system of the soft gripper when (a) the valve operation for inflation turns on; (b) the valve operation for deflation turns on.}
      \label{fig::airflowgripper}
   \end{figure}
   
As the soft drone approaches the target object, a deflation command is transmitted from the ground control station to the flight controller and subsequently to the ESP32-S3 microcontroller. This command activates the deflation airflow through the valves and initiates vacuuming with the air pump. The deflation process allows the soft gripper to open, release, and land, while the inflation process activates the grasping of the gripper by complete closing. During grasping, deflation continues until an inflation command is received, triggering the gripper to close. Inflation follows a similar procedure to deflation. The microcontroller receives three ranges of PWM inputs to control the behavior of the soft gripper: staying at rest, inflation, and deflation. The airflow patterns during inflation and deflation are depicted in Fig. \ref{fig::airflowgripper}. Both valves are turned off to prevent airflow when the gripper is initially opened. Only the relevant valve is activated during inflation or deflation, while the other valve remains off to ensure proper airflow direction. In Fig. \ref{fig::airflowgripper}, the blue solid block and the red solid block of the air pump represent the inflation and deflation function of the pump that is mainly used in the relevant airflow, respectively. Also, the blue and the red airflows indicate the inflation and the deflation airflows of the gripper, respectively.

\subsection{Pressure Regulation}
\label{subsubsec::airpumpregulation}

\begin{figure}[t]
    \centering
    \includegraphics[scale = 0.21]{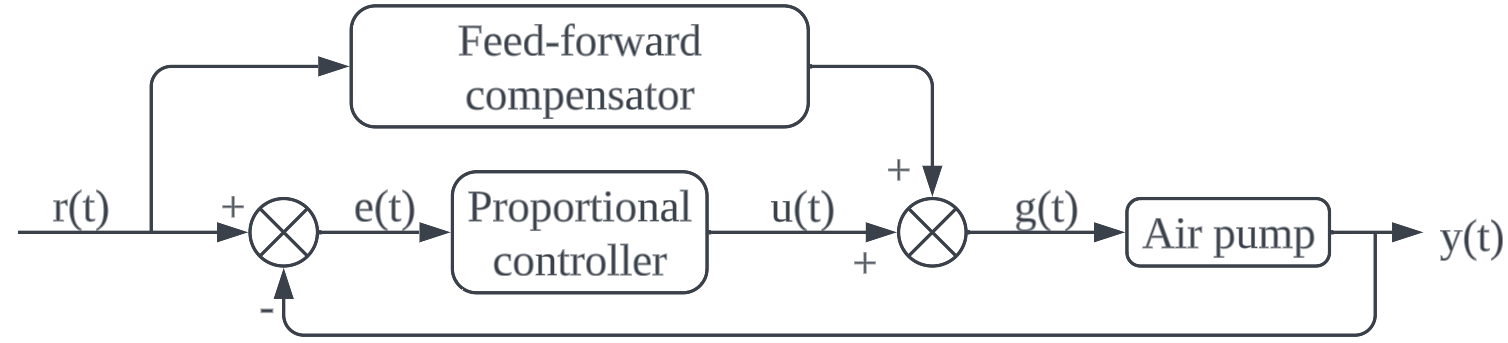}
    \caption{Block diagram of feed-forward proportional control for the soft gripper.}
    \label{fig::feedforward_p_control_gripper}
\end{figure}

The pressure regulation of the soft gripper employs a feed-forward proportional control strategy, as presented in Fig. \ref{fig::feedforward_p_control_gripper}. This approach combines feed-forward and proportional control techniques to enhance performance by anticipating the impact of input pressure changes on the output pressure. Due to the limited performance of the lightweight air pump and the unseal soft gripper system, which enables easy swapping between two configurations, a slight air leakage can be predicted as a disturbance of the pressure regulation. By integrating these control strategies, the gripper achieves faster response times and improved stability in regulating the air pump. Notably, the feed-forward proportional control only sets the PWM signal for the air pump. At the same time, the PWM command of the flight controller directly determines the digital signals for activating the valves. This valve control strategy prevents excessive vibration caused by rapid changes in airflow direction.

The feed-forward proportional controller in Fig. \ref{fig::feedforward_p_control_gripper} can be computed by solving the following equations:

\begin{equation}\label{eqn::p_error}
    e(t) = r(t) - y(t)
\end{equation}
\begin{equation}\label{eqn::p_output}
    u(t) = K_p \times e(t) + p_{min}
\end{equation}
\begin{equation}\label{eqn::p_gain}
    K_p = \dfrac{p_{max} -p_{min}}{r(t)}
\end{equation}
\begin{equation}\label{eqn::ff_inflation}
    f_{in} = 0.8
\end{equation}
\begin{equation}\label{eqn::ff_deflation}
    f_{de} = \dfrac{1}{r(t)}
\end{equation}
\begin{equation}\label{eqn::ff_p_control}
    g(t) = \begin{cases}
        u(t) + f_{in} \times y(t) &\text{if inflation}\\
        u(t) + f_{de} \times y(t)  &\text{if deflation}
        \end{cases}
\end{equation}

\noindent where the minimum PWM duty cycle of the air pump, $p_{min}$, is an adjustable parameter to optimize the performance according to the desired pressure. If $p_{min}$ is too low, the response time of the system will be longer. The maximum PWM duty cycle of the air pump, $p_{max}$, is fixed at the 8-bit capacity of the microcontroller of the gripper. Referring to the experimental results, we also observe the optimal feed-forward component for inflation, $f_{in}$, and deflation $f_{de}$ to speed up the inflation and deflation time of the soft gripper. 
Further details and explanations of the abovementioned variables can be found in Table \ref{tab::ff_p_control}.

\begin{table}[t]
 \begin{center}
 \renewcommand{\arraystretch}{1.3}
 \caption{Parameters in air pump regulation.}
 	\begin{tabular}{c c}
		\hline 
            Variables & Definitions \\
		\hline 
		  $e(t)$ & Pressure error in $kPa$\\
            $r(t)$ & Set (desired) pressure\\
            $y(t)$ & Current pressure\\
            $u(t)$ & Output by Proportional controller\\
            $Kp$ & Proportional gain\\
            $p_{max}$ & Max. PWM duty cycle of air pump $(100\%)$\\
            $p_{min}$ &  Min. PWM duty cycle of air pump $
                \begin{cases}
                $86\%$ &\text{if inflation}\\
                $63\%$ &\text{if deflation}
                \end{cases}
                $\\
            $f_{in}$ & Feed-forward component for inflation\\
            $f_{de}$ & Feed-forward component for deflation\\
            $g(t)$ & Output by Feed-forward proportional controller\\
        \hline
	\end{tabular}
    \renewcommand{\arraystretch}{1}
	
	\label{tab::ff_p_control}
  \end{center}
\end{table}

\begin{figure}[t]
    \centering
    \includegraphics[scale = 0.24]{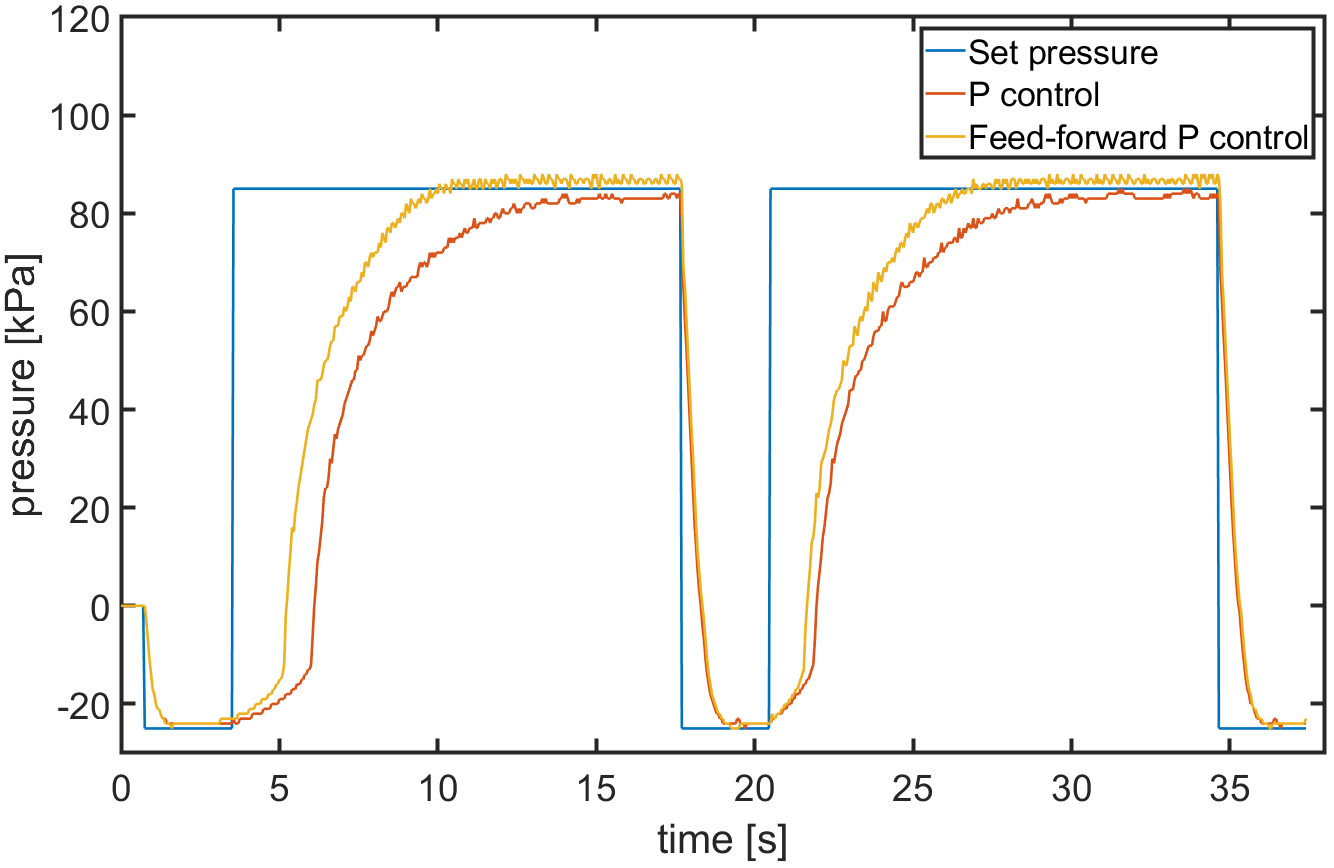}
    \caption{Comparison between feed-forward proportional controller and proportional controller of the soft gripper.}
    \label{fig::feed_forward_P_control}
\end{figure}

The soft gripper focuses on achieving maximum openness and a secure grasp by pressure regulation. After considering factors such as the air chamber capacity, inflation time, performance of the lightweight air pump, and the results of the feed-forward proportional controller, an inflation pressure of 85 kPa was determined as optimal for achieving the desired level of openness. Also, a deflation pressure of -25 kPa was chosen to ensure rapid and effective closure. We also observe the optimal $f_{in}$ and $f_{de}$ to speed up the inflation and deflation time of the soft gripper. The step response graph in Fig. \ref{fig::feed_forward_P_control} illustrates the pressure regulation performance with a set pressure signal (set pressure in 0 kPa, -25 kPa, and +85 kPa) by the feed-forward proportional controller and the proportional controller. Although the deflation performance for both controllers is similar, the inflation time of the feed-forward proportional controller is much faster than that of the proportional controller. These results show that the feed-forward component anticipates set pressure changes, facilitating stable current pressure and reduced steady-state error. Also, in the feed-forward proportional control, the desired inflation pressure can be reached faster, compared with the proportional control only, but the desired deflation performance of both control methods differs insignificantly. Note that the soft gripper starts closing at 58 kPa and takes approximately 5 seconds to reach the proposed inflation pressure from deflation. 

\section{EXPERIMENTAL RESULTS AND DISCUSSION}
\label{sec::resultsanddiscussion}
\subsection{Experiment Implementation}
\subsubsection{Aerial Grasping Test}
\label{subsubsec::aerial_grasping_test}
During the indoor autonomous grasping tests performed with the proposed soft gripper installed under a conventional drone (with an F330 airframe), it is observed that the mass of the object being grasped significantly influences the grasping process. Objects that are too light (less than approximately 70 g) are susceptible to being blown away by the airflow generated by the SAV before aerial grasping. Therefore, using objects with a mass greater than 70 g is recommended. In the aerial grasping task, the target object was a water bottle (including loads) weighing 75 g, while the proposed SAV weighed 808 g. The water bottle was intentionally emptied to eliminate any non-static mass caused by liquid flow during grasping. To prevent the water bottle from being blown away by the airflow of the SAV, additional loads were added to increase its weight. 

The flight tests were conducted under a VICON motion capture system, which provided real-time ground truth data to the ground station computer to assist the flight controller in tracking the target item. The proposed SAV uses a nonlinear model predictive controller (NMPC) \cite{jiang2022neural, li2018development} for minimizing tracking error and conducting aerial grasping. Despite slight tracking errors during grasping, the softness of the soft gripper compensated for them, offering sufficient grasping tolerance. Thus, the water bottle was securely grasped without being damaged. The flight test aims to demonstrate the aerial grasping ability of the proposed modular pneumatic soft gripper, so the tracking error due to the change in the dynamic model of the SAV after grasping is beyond the scope of this paper and neglected. This paper does not include the details of the NMPC for the SAV, as our focus is on highlighting the contributions of the modular pneumatic soft gripper in aerial grasping.

\subsubsection{Static Grasping Test}
\label{subsubsec::soft_grasping_test}
To further evaluate the grasping performance of the SAV, a static grasping test aimed to assess both the maximum load capacity of the soft gripper and its ability to grasp objects with various base shapes. The test involved evaluating the performance of the gripper under different load conditions and its ability to grasp objects with diverse base configurations securely.

\begin{figure}[t]
    \centering
    \includegraphics[scale = 0.25]{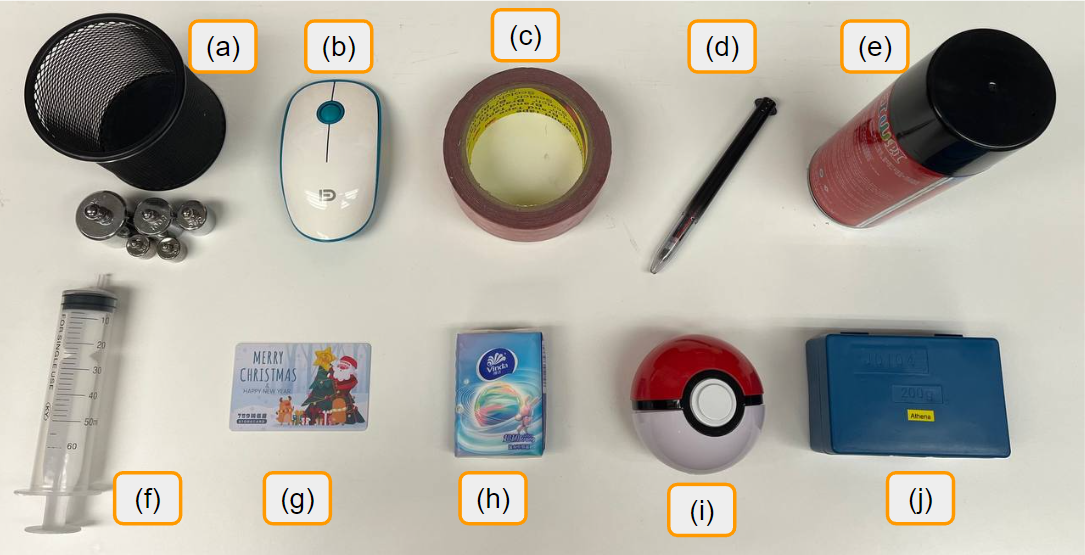}
    \caption{Tested Object Set: (a) Pen Holder with Loads, (b) Computer Mouse, (c) Double-sided Tape, (d) Pen, (e) Spray Paint, (f) Syringe, (g) Membership Card, (h) Pocket Tissue Paper, (i) Spherical Container with Loads, (j) Plastic Box.}
    \label{fig::target_objects}
\end{figure}

In this experiment, the SAV was put on an aluminum rack, and the servo tester provided the inflation or deflation PWM input signal to the microcontroller of the gripper, making it independent of the drone system without receiving inputs from the flight controller or ground control station. Target objects were manually placed under the center of the gripper by human hands. To test the grasping tolerance of the gripper, the positions of the centroid of the objects did not require a precise alignment of the centroid of the gripper before grasping. 10 graspings were conducted for each specimen. The 10 objects in the grasping test are shown in Fig. \ref{fig::target_objects}. Successful grasping trials required the gripper to hold the target object for at least 30 seconds. To ensure a reliable grip, human hands intentionally moved and gently shook the SAV during the test.

\subsubsection{Soft Landing Test}
\label{subsubsec::soft_landing_test}
The soft gripper establishes its landing ability by keeping a full opening as a soft landing gear to stabilize the pose of the SAV and dampen the impact forces during landing. The proposed soft gripper replacing the rigid landing gear can increase the efficiency of the aerial manipulation of the SAV without burdening its payload capability. To assess the landing ability of the SAV, the SAV first conducted a simple landing test, which requested the SAV to land on the ground. Then, the tilt landing test of the SAV was also handled by landing it on a slope with 10$^{\circ}$ inclined angles.

\subsubsection{Payload Test of the SAV}
\label{subsubsec::soft_payload_test}
The maximum payload of the SAV with two configurations of its soft gripper was discovered by grasping different objects while hovering. To focus on the payload capability, the aerial grasping ability of the SAV was ignored in this test by grasping the target items before taking off.
\subsection{Results and Discussion}
\subsubsection{Soft Grasping Performance}
\label{subsubsec::soft_grasp_performance}

\begin{figure}[t]
    \centering
    \includegraphics[scale = 0.3]{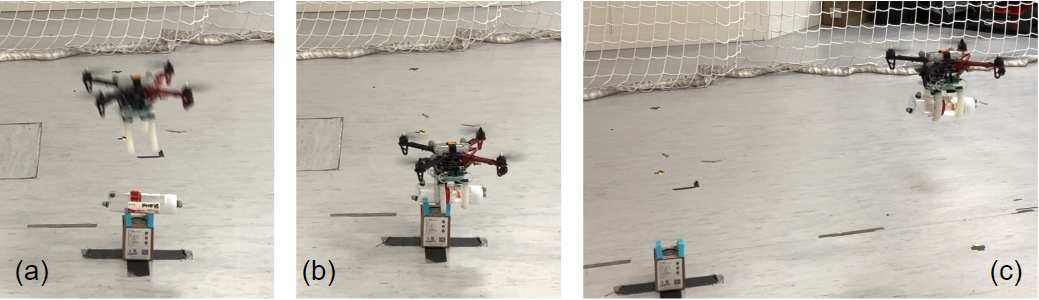}
    \caption{Aerial grasping process with the H-base soft gripper: (a) Approaching the target, (b) Starting grasping, (c) Picking up the object.}
    \label{fig::aerial_grasping}
\end{figure}

Figure \ref{fig::aerial_grasping} illustrates a successful aerial grasping demonstration performed by the proposed H-base soft gripper. Despite minor tracking errors causing the SAV to miss the desired grasping position of the empty water bottle, the inherent softness of the soft gripper allows it to compensate for such errors and provide ample grasping tolerance. As a result, the bottle could be securely grasped without sustaining any damage.

\begin{figure}[t]
    \centering
    \includegraphics[scale = 0.35]{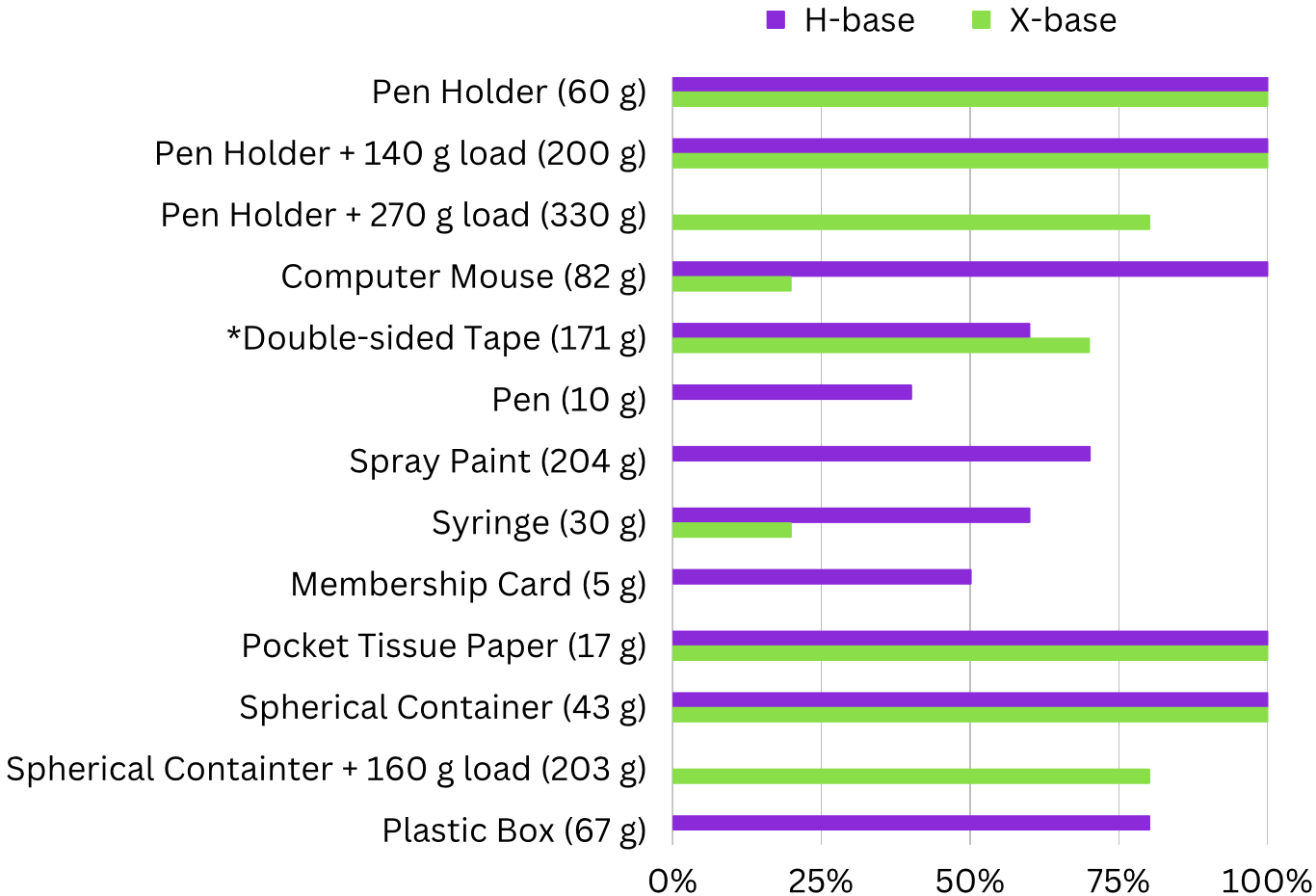}
    \caption{Grasping success rate results of 2 configurations. \\ *The X-base gripper can hang and wrap the tape, while the H-base gripper always wraps the tape.}
    \label{fig::grasp_success_rate}
\end{figure}

\begin{figure}[t]
    \centering
    \includegraphics[scale = 0.35]{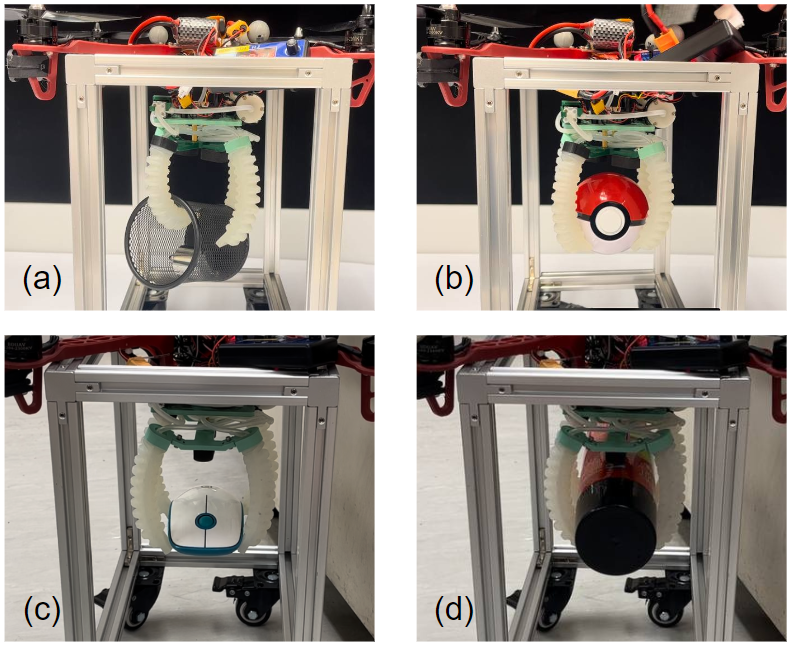}
    \caption{Examples of grasping tests: Grasping (a) a pen holder with 270 g loads and (b) a spherical container with 160 g loads by an X-base gripper; Grasping (c) a computer mouse and (d) a can of spray paint with an H-base gripper.}
    \label{fig::grasp_example}
\end{figure}

From Fig. \ref{fig::grasp_success_rate}, the soft gripper with both two configurations can grasp the pen holder and the pocket tissue paper with a 100\% success rate. The 4-tip soft gripper with an H-base is designed for grasping cylindrical objects and offers increased gripping force through the wrapping motion of its two pairs of 2-tip fingers. The H-base gripper can grasp cylindrical objects up to 200 g with a 100\% success rate, but it has an 80\% success rate for grasping a can of spray paint due to the non-static center of gravity. It cannot handle items more miniature than the distance between its finger pairs, so it is less effective with spherical objects. To address these limitations, a proposed X-base allows all four fingers to align with the centroid of objects to grasp spherical items. It can also grasp the pen holder with loads up to 330 g because one of the soft fingers can hang the pen holder with the support of the other three soft fingers. Nonetheless, the X-base does not provide the same wrapping force for cylindrical objects as the H-base. Thus, X-base and H-base serve distinct purposes. Some examples of this static grasping test are illustrated in Fig. \ref{fig::grasp_example}.

\subsubsection{Soft Landing Performance}
\label{sec::soft_landing_performance}

\begin{figure}[t]
    \centering
    \includegraphics[scale = 0.35]{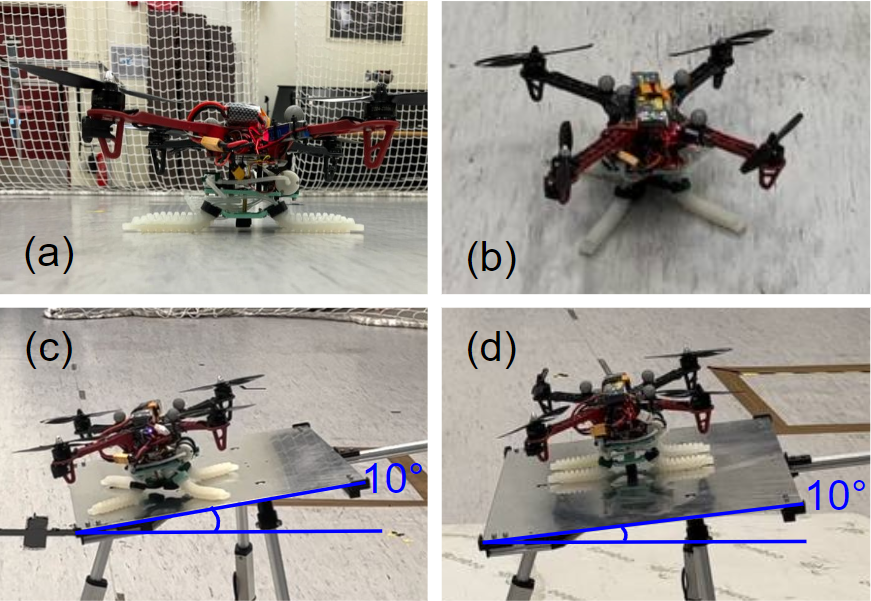}
    \caption{Simple landing test with (a) the H-base soft gripper and (b) the X-base soft gripper, and tilt landing test with the (c) X-base and (d) H-base soft gripper.}
    \label{fig::soft_landing_gear}
\end{figure}

\begin{table}[t]
 \begin{center}
 \renewcommand{\arraystretch}{1.3}
 \caption{Comparison of SAV landing performance on ground and tilt platform.}
        \begin{tabular}{@{}>{\centering\arraybackslash}m{1.5cm}|>{\centering\arraybackslash}m{2.5cm}>{\centering\arraybackslash}m{2.5cm}@{}}
        \hline 
            Configurations & Successes (ground) & Successes (tilt)\\
		\hline 
		  X-base & $10/10 (100\%)$ & $6/10 (60\%)$ \\
            H-base & $10/10 (100\%)$ & $10/10 (100\%)$\\
        \hline
	\end{tabular}
    \renewcommand{\arraystretch}{1}
	\label{tab::landing_performance}
  \end{center}
\end{table}

In Fig. \ref{fig::soft_landing_gear}(a) and Fig. \ref{fig::soft_landing_gear}(b), with the modular soft gripper, this novel SAV can take off and land on the ground without the need for traditional rigid landing gear. Since the landing pressure is equivalent to the deflation pressure used during grasping, the soft gripper can remain fully opened, serving as a landing gear to stabilize the pose of the SAV during takeoff and effectively dampen impact forces upon landing. Besides, Fig. \ref{fig::soft_landing_gear}(c) and Fig. \ref{fig::soft_landing_gear}(d) present the successful tilt landing of the SAV (with its two configurations of the soft gripper) on a 10$^{\circ}$ inclined platform. According to Table \ref{tab::landing_performance}, the H-base soft gripper performs better for the tilt landing as it can always be flattened, while the X-base soft gripper sometimes may only have three fingers to be flattened and even jump out from the platform due to the impact forces. These tilt landing results show that the H-base soft gripper can dampen the impact forces of the soft fingers more since the fingers have more contact areas with the tilt platform. Thus, the modular soft gripper can successfully replace conventional rigid landing gear, even for tilt landing.

\subsubsection{Payload Ability Performance}
\label{subsubsec::payload_ability}

\begin{figure}[t]
    \centering
    \includegraphics[scale = 0.23]{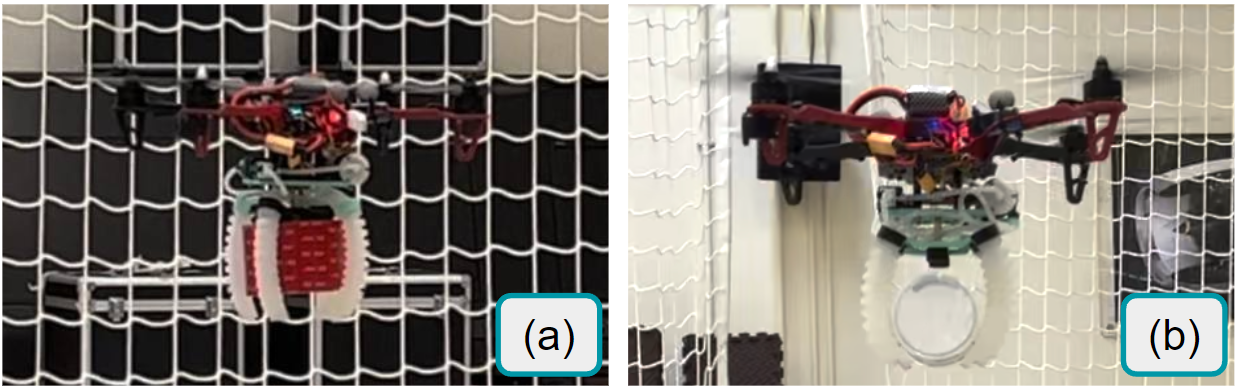}
    \caption{Payload test of the SAV with the X-base (a) and (b) H-base soft gripper.}
    \label{fig::paylaod}
\end{figure}

In Fig. \ref{fig::paylaod}, the SAV demonstrates its successful grasping of a 217 g double-sided tape using the X-base soft gripper and a 217 g plastic cylindrical container using the H-base soft gripper. Both payload tests were conducted for over 30 seconds. This payload ability is higher than those in some previous research, such as 106 g foam target in \cite{fishman2021dynamic}, 100 g payload in \cite{pingaerial}, and 150 g packet in \cite{sarkar2022development}. The focus of this test was to evaluate the payload capacity of the SAV, with the aerial grasping capability intentionally excluded. 

\section{CONCLUSION}
\label{sec::conclusion}
We have presented the design and control of a modular pneumatic soft gripper equipped under a traditional quadrotor (556 g) as a novel SAV for autonomous aerial grasping tasks and landing. The main objective of this soft gripper design is to maintain a lightweight structure, thereby minimizing the impact on the flight capabilities of the SAV. Therefore,  the proposed modular pneumatic soft gripper weighs less than 260 g. Pressure control for the modular gripper utilizes a feed-forward proportional controller to improve pressure regulation. Experimental results from grasping tests reveal that the total contact areas of the gripper on the target object influence the grasping force. To accommodate objects of different shapes, we have explored two base configurations for the modular soft gripper. The H-base 4-tip gripper is suitable for cylindrical or rectangular objects, while the X-base 4-tip gripper is more appropriate for spherical or rounded objects. 

This paper also demonstrates a successful indoor autonomous aerial grasping task. The SAV can grasp a 217 g payload in mid-air, which signifies a satisfactory payload-to-weight ratio. Furthermore, the soft gripper efficiently replaces rigid landing gear, facilitating competent landings on both the ground and a 10$^{\circ}$ inclined platform. These findings contribute to developing autonomous aerial grasping capabilities in soft robotic systems. We also discover two future challenges, including adapting to the dynamic model changes after grasping and enhancing the grasping ability on thin and small objects.



\bibliographystyle{IEEEtran}
\bibliography{references}

\end{document}